# A Comprehensive Study on Fine-Tuning Large Language Models for Medical Question Answering Using Classification Models and Comparative Analysis

Ayşegül Uçar[1*], Soumik Nayak[2], Anunak Roy[3], Burak Taşcı[4], Gülay Taşcı[5]

[1]Department of Mechatronics Engineering, Engineering Faculty, Fırat University, Elazığ 23119, Turkey

[2]Department of Mechanical Engineering, Indian Institute of Technology Kharagpur, Kharagpur, India

[3]Department of Electrical Engineering, Indian Institute of Technology Kharagpur, Kharagpur, India

[4]Vocational School of Technical Sciences, Fırat University, Elazığ 23119, Turkey

[5]Department of Psychiatry, Elazig Fethi Sekin City Hospital, Elazig, Turkey

Corresponding author [1*]Email: agulucar@firat.edu.tr

**Abstract**

This paper presents the overview of the development and fine-tuning of large language models (LLMs) designed specifically for answering medical questions. We are mainly improving the accuracy and efficiency of providing reliable answers to medical queries. In our approach, we have two stages, prediction of a specific label for the received medical question and then providing a predefined answer for this label. Various models such as RoBERTa and BERT were examined and evaluated based on their ability. The models are trained using the datasets derived from 6,800 samples that were scraped from Healthline. com with additional synthetic data. For evaluation, we conducted a comparative study using 5-fold cross-validation. For accessing performance we used metrics like, accuracy, precision, recall, and F1 score and also recorded the training time. The performance of the models was evaluated using 5-fold cross-validation. The LoRA Roberta-large model achieved an accuracy of 78.47%, precision of 72.91%, recall of 76.95%, and an F1 score of 73.56%. The Roberta-base model demonstrated high performance with an accuracy of 99.87%, precision of 99.81%, recall of 99.86%, and an F1 score of 99.82%. The Bert Uncased model showed strong results with an accuracy of 95.85%, precision of 94.42%, recall of 95.58%, and an F1 score of 94.72%. Lastly, the Bert Large Uncased model achieved the highest performance, with an accuracy, precision, recall, and F1 score of 100%. The results obtained have helped indicate the capability of the models in classifying the medical questions and generating accurate answers in the prescription of improved health-related AI solutions.

**Keywords:** Medical Question Answering; Large Language Models; RoBERTa; BERT; Healthline Dataset; Model Evaluation Metrics

## 1. Introduction

Recent discoveries in Artificial intelligence have impacted various areas like healthcare in a big way. An example of this is the develoloptment of large language models (LLMs)[1, 2], that are capable of interpreting and generating text in a human-like manner. It is used in probable solutions to health related questions and identifying certain diseases based on the symptoms displayed. This work is therefore designed to take advantage of the functionalities offered by LLMs in developing a sound system for answering disease related questions[2]. It directs the adjustment of existing models like Roberta and BERT for the enhancement of medical QA models[3]. Additionally we focus on the possibilities of simplifying the training process of these models with the help of Low-Rank Adaptation (LoRA)[4]. The proposed approach is based on first classifying medical questions on predefined classes . It then enables each question to get an appropriate response appropriate to the respective category of the question.

For achieving this, we used a data set from the HealifyAI-LLM based Healthcare System project which consists of 6,800 samples scraped from healthline.com with additional synthetic samples[5]. This dataset is chosen to give as many medical questions as it is possible, and give detailed answers to such questions. For this experiment the first dataset was divided into the training, validation and the tested set. In our study, we are fine-tuning four models namely LoRA Roberta-large; Roberta-base; Bert Uncased; and Bert Large Uncased. To validate the performance, we use a 5-fold cross-validation approach. These models were assessed in terms of key metrics like accuracy, precision, recall and F1 score, across various folds and epochs. Such an evaluation framework covers all significant aspects of the models and guarantees that their effectiveness will be assessed properly. Further, we also compare the training time of these models. The experiments were performed on the T4 GPU which is capable enough to perform the training and evaluation of models. Hence, by comparing these models, we expect to find the model best suited for medical question answering. The information gained from this study might help enhance the creation of better and more sophisticated AI systems in the sphere of healthcare, which in turn will improve patient care and their access to credible medical information. The use of a high-performance GPU makes sure that the models are trained effectively and time is reduced and can be used on more iterations. In summary, this research aims to address the challenges of medical question answering by leveraging the power of LLMs and innovative fine-tuning techniques. The contribution of this study can be considered as rather valuable for the development of AI in the healthcare context and can offer the tools to help professionals increase the rates of patient recovery. In the subsequent sections, we will detail the methodology, results, and conclusions of our study. We will also make some recommendations on the basis of our research and potential future directions for research in this field.

## 2. Related Works

Some of the domains that have greatly benefited from LLMs in healthcare include question answering and disease prediction. This section provides an overview of the literature that underpins and situates this research.

Raghu Subramanian et al[6], evaluated the role of large language models (LLMs) in health communication. Their study focused on the process of transforming hospital discharge summaries into more comprehensible formats for patients using the GPT-4 model. Among 100 discharge summaries analyzed, 54% were successfully transformed, while 18 summaries revealed potential safety risks. Although the approach enhanced readability and comprehensibility, safety risks and challenges in integrating the model into clinical workflows were identified as limitations. Stade et al.[7], proposed a roadmap for the responsible development and evaluation of integrating LLMs into psychotherapy. Their study explored the use of models like GPT-4 in psychotherapeutic processes and highlighted their potential to address accessibility challenges and enhance personalized treatment options. The study provided detailed insights into the multi-dimensional integration stages of these models and the associated ethical responsibilities. However, it emphasized that safety concerns, biases, and ethical constraints remain critical challenges when using such models in complex, high-risk domains like psychotherapy. Lawrence et al.[8], examined the opportunities and risks of LLMs in the field of mental health. Their study summarized the applicability of these models in mental health education, assessment, and intervention, highlighting key opportunities to create positive impacts in these areas. However, careful development, testing, and implementation strategies were recommended to ensure ethical use. The study specifically underscored the importance of adapting LLMs for mental health applications, promoting mental health equity, and adhering to ethical standards. Raza et al.[9], investigated the applicability of generative artificial intelligence (AI) and LLMs in healthcare. The study proposed a framework for evaluating models that leverage electronic medical records (EMRs), focusing on criteria such as predictive performance, data labeling, ease of implementation, multimodality, and human-AI interactions. Limitations such as lack of generalizability and concerns regarding data privacy were identified as significant challenges for current models. Tian et al.[10], discussed the opportunities and challenges of ChatGPT and other LLMs in biomedical and healthcare applications. Their study emphasized applications such as information retrieval, question answering, text summarization, and medical education, noting significant advancements over existing methods. However, risks related to data privacy and response accuracy were highlighted as key limitations. The study stressed the need to develop ethical and legal frameworks to ensure the effective and responsible use of these technologies.

**Comparison of LLMs:**

The comparative analysis of different LLMs like RoBERTa and BERT has helped in getting insights about their strengths and weaknesses . In a direct comparison, Liu et al.[11] showed that due to the more extensive pretraining and optimised hyperparameters RoBERTa achieves better results in many NLP tasks as compared to BERT. These have particularly been used in selecting the right models for specialised tasks such as medical question answering. Subsequent research has also investigated the approach of model compression by other authors like Lan et al. [12] with ALBERT to establish the ways of achieving the optimal trade off between the model performance and resource utilisation

**Low-Rank Adaptation (LoRA):**

The work of Hu et al. [4] in developing LoRA provided the foundation for future efforts to make large models more accessible for fine-tuning on certain tasks. Therefore, LoRA minimises the number of trainable parameters using low-rank matrices thus enabling efficient adaptation of LLMs without compromising their performance. This method has proved to be particularly useful in situations where computational resources can be a challenge. Subsequent work by Ziegler et al.[13] has furthered this line of work, showing that LoRA can be used across many different models and tasks, including medical applications.

**Medical Question Classification:**

The identification of placing medical questions into predefined categories is one of the most important steps in order to provide users with adequate results and information. Namely, Lee et al.[14] proposed a transformer-based method for classifying medical queries with high accuracy and recall,setting a benchmark for future work in this domain(Biobert). Therefore, this research targets the accurate classification of questions as a basic requisite of medical question answering systems.

### 3. Contribution

IIn this study, significant contributions were made in several key areas such as Model Fine-Tuning and Comparative Analysis

**Model Fine-Tuning and Comparative Analysis**:

Tuned four models which include LoRA Roberta, Roberta, Bert and large Bert Uncased for the task of medical question answering. . All the 5 fold cross validation on the two models were done on 10 epochs each.

Conducted an assessment with the purpose of comparing these models in the context of the measures that can be applied to assess the performance of these models for medical use for accuracy, precision, recall and F1 score.

**Innovative Use of LoRA**:

Proposed improving the fine-tuned models such as Roberta and BERT using Low-Rank Adaptation (LoRA) to show that accuracy could be maintained even after reducing the computational cost

**Evaluation and Metrics Analysis**:

During the training process, we had to monitor and compare the performance of the model from one fold to another and from one epoch to another and therefore had to monitor, record, and visualise the performance characteristics of the model.

Comparing training times for each model in order to get an accurate picture of the models' performance when it comes to computational cost in a realistic, large-scale setting.

Thus, this research contributes to the literature of using LLMs in the healthcare domain and demonstrates the role of AI in enhancing the quality of care and providing patients with credible medical data.

## 4. Datasets

For this study, we employed two datasets from the HealifyAI LLM–based–Healthcare–System project of Tanvir Ishraq available on GitHub[5]. These datasets are to improve the ability of large language models (LLMs) in the healthcare field specifically. Here, we give a brief overview of the two datasets, from where they were obtained and the pre-processing that has been done.

The main dataset consists of 6,800 samples which have been systematically constructed from scratch. Data collection was done through scraping information relating to health from the website http://www.healthline.com. com, a reliable source of medical information. To augment the volume of the given dataset and enhance the utility of the tool for the user, more samples were created with the help of a Python program. This process made sure it could answer as many questioning methodologies as possible and in detail and with precision.

The second data set offers the specific answers that correspond to particular labels associated with different diseases. This dataset was also generated by data scraping from the website; healthline. com and arrange it in a manner to correspond to the label I am predicting from the questions I have asked.

Both datasets used in this study were developed by Tanvir Ishraq as part of the HealifyAI project. The data was sourced from the Healthline web page and was extended with newly generated samples to enhance its variety and representativeness. In preparing the datasets for training and evaluation, several preprocessing steps were applied, including data cleaning to remove unnecessary or redundant information, normalization to establish a consistent structural pattern, tokenization to convert text into entities suitable for large language models (LLMs), and labeling to categorize the data based on disease-related questions. The study focuses on utilizing the primary dataset for all observations and experiments, ensuring that the results and conclusions are based on a unified and comprehensive data set explicitly designed for medical question classification and relevant answer generation. Using these curated datasets, the objective is to train an accurate and reliable LLM capable of addressing various disease-related inquiries. The primary dataset was divided into training and validation sets to evaluate model performance. Specifically, 70% of the data (4,760 samples) was used as the training set, while the remaining 30% (1,020 samples) was allocated to validation for model tuning and hyperparameter adjustments. The primary dataset is organized in a tabular format with columns for "Disease," "Question," and "Label." The "Disease" column indicates the type of disease, the "Question" column includes medical inquiries, and the "Label" column corresponds to each question's specific label. The secondary dataset also follows a tabular format, including "Disease," "Label," and "Answer" columns, where "Disease" categorizes the type of disease, "Label" represents each disease-specific label, and "Answer" provides a detailed response corresponding to the label. Ethical standards and privacy issues were considered while using the datasets.Since the data was scraped from publicly available information on healthline.com , there are no direct privacy issues.

The primary and secondary datasets used in this study are organized to facilitate medical question classification and relevant answer generation. Table 1 presents sample entries from the primary dataset, comprising disease-related questions along with their corresponding labels. Table 2 provides a glimpse of the secondary dataset, which includes disease categories, labels, and detailed answers that support model training.

**Table 1.** Primary Dataset[5]

| Disease | Question | Label |
|---|---|---|
| diabetes | What is diabetes? | diabetes definition |
| diabetes | Tell me about diabetes? | diabetes definition |
| diabetes | What kind of disease is diabetes? | diabetes definition |
| diabetes | Can you elaborate on diabetes? | diabetes definition |
| diabetes | What can you tell me about diabetes? | diabetes definition |

**Table 2.** Secondary Dataset[5]

| Disease | Label | Answer |
|---|---|---|
| diabetes | diabetes definition | Diabetes mellitus is a metabolic disease that causes high blood sugar. |
| diabetes | diabetes symptoms | Diabetes symptoms are caused by rising blood sugar levels. Common symptoms include increased... |
| diabetes | diabetes causes | Different causes are associated with each type of diabetes. Type 1 diabetes is an autoimmune... |
| diabetes | diabetes risks | Certain factors increase your risk for diabetes. Type 1 is thought to be caused by genetic s... |
| diabetes | diabetes | High blood sug... damages organs and tissues throughout your body. The... |

## 5. Research Methods

In this section, we present the methodology adopted for fine-tuning large language models (LLMs) for the classification task. The approach involved using four different models, with their performance evaluated through rigorous cross-validation and metric analysis.

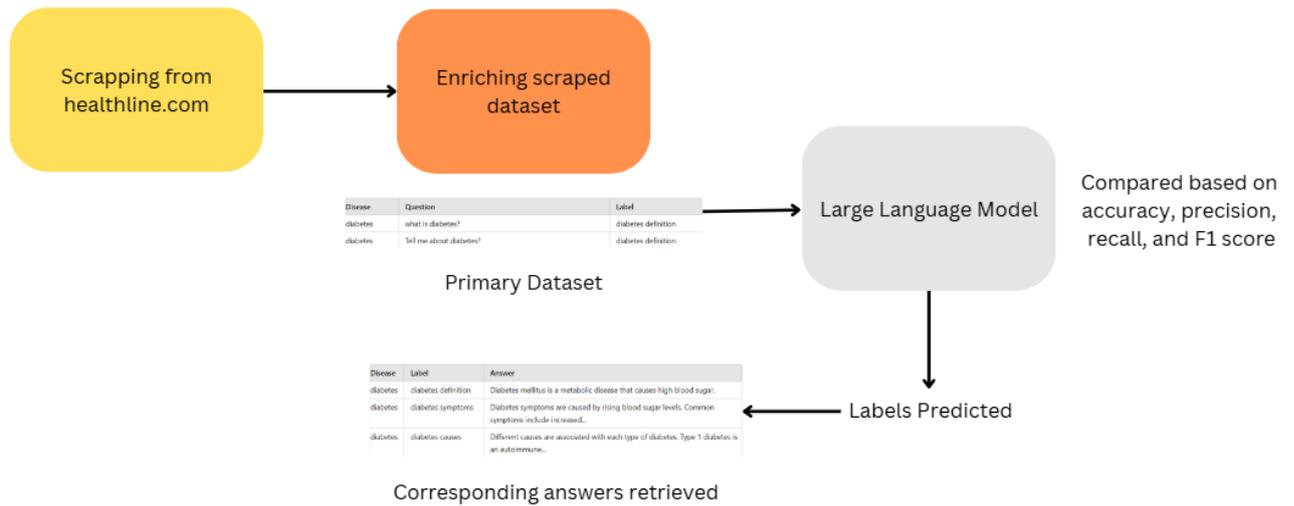

**Figure 1** Block diagram of the proposed method

Step 1: Models Used In this study, four models were fine-tuned, each selected for its specific capabilities. The models included: (1) LoRA Roberta-large, which employs Low-Rank Adaptation to reduce trainable parameters while maintaining high accuracy, (2) Roberta-base, a smaller yet practical version of Roberta, (3) Bert Uncased, a model designed to convert text to lowercase and remove accent characters, and (4) Bert Large Uncased, a higher-capacity version of the BERT model containing more parameters than the base variant.

Step 2: Training Methodology For model fine-tuning, we employed a 5 fold cross validation strategy, which was appropriate given the size of the dataset. In this process, the dataset was divided into five parts, with four parts used for training and the remaining one for validation. This process was repeated five times, with each part used once as the validation set. This methodology helps prevent overfitting and provides a comprehensive assessment of the model's general performance.

Step 3: Training Details Training was conducted for 10 epochs for each fold to allow the models to sufficiently learn from the data without overfitting. The performance evaluation metrics included accuracy, precision, recall, and F1 score, calculated for each fold and epoch. The average values across all folds were used for the final assessment. In addition to metric calculation, graphical analysis was conducted to plot these metrics across each fold and epoch, facilitating the identification of potential issues like overfitting or underfitting. Training times were also recorded for each model to compare their computational efficiency.

Step 4: Evaluation Metrics The models were evaluated using several metrics: (1) Accuracy, which represents the ratio of correctly predicted instances to the total instances, (2) Precision, indicating the ratio of correctly predicted positive observations to the total predicted positives, (3) Recall, defined as the ratio of correctly predicted positive observations to all observations in the actual class, and (4) F1 Score, which is the weighted average of precision and recall, providing a balance between these two metrics. These metrics were derived using the fastai.metrics module from the Fastai library, a high-level deep learning library built on top of PyTorch. Fastai facilitates the development of deep learning models by offering pre-built functions, classes, and modules that simplify tasks such as training neural networks, handling data, and implementing state-of-the-art models.

**Experimental Setup:**

The experimental setup for this study involved conducting all experiments on a Google Colab T4 GPU, which provided an accessible environment for training and evaluation. During the experiments, hyperparameters such as learning rate, batch size, and optimizer were carefully examined and tuned for each model to achieve the best possible performance. Detailed records of the training progress were kept, capturing performance data across different folds and epochs to precisely monitor and evaluate the effectiveness of each model.

### 6. Results

In total four models were fine-tuned for this text classification task:

1) LoRA Roberta-large
2) Roberta-base
3) Bert Uncased
4) Bert Large Uncased

All these models were trained using 5-fold cross validation, each fold being a total of 10 epochs.

LoRA Roberta-large

The LoRA Roberta-large model was evaluated using a 5-fold cross-validation approach, and the average performance metrics were as follows: The model achieved an accuracy of 78.47%, a precision of 72.91%, a recall of 76.95%, and an F1 score of 73.56%.

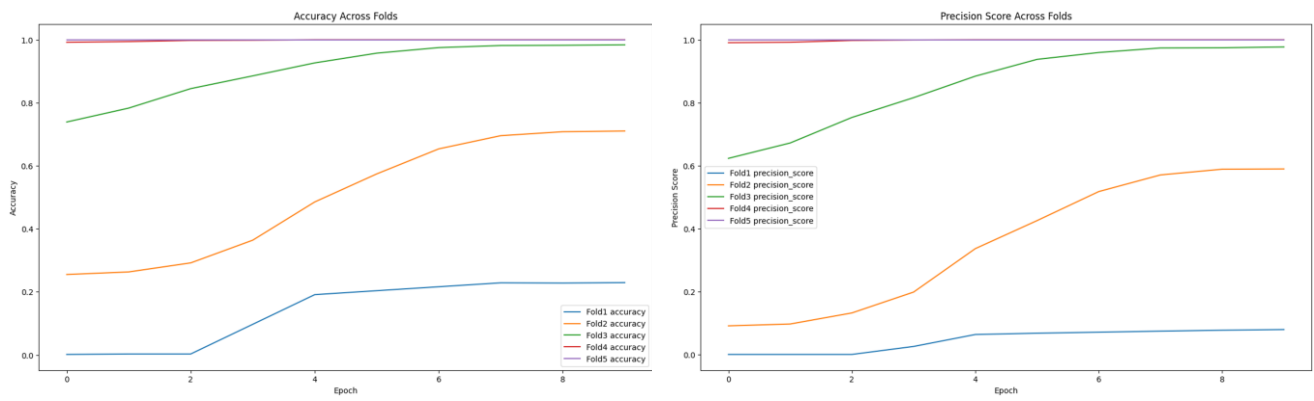

a  b

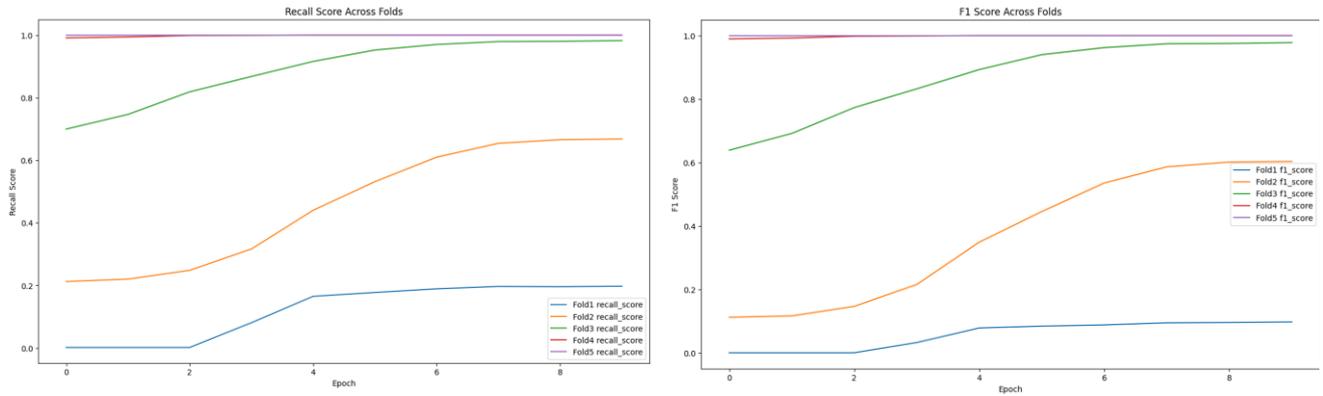

*Figure 2* LoRA Roberta-large graphics a:accuracy b:Precision c: Recall d:F1 score

These are plots of cross-validated performance metrics on the text classification downstream task for a LoRA-tuned RoBERTa-large model trained and evaluated across 10 epochs, with results reported per fold. Folds 4 and 5 almost always score near perfectly, with slower improvement in performance in the initial few folds. While folds 1 and 2 lag slightly, fold 3 trends towards the top performers by the final epochs while ultimately ending up at a similar level. The model seems to learn fast with most of the gains being achieved in the first 4 - 6 epochs before plateauing.

Roberta-base

The Roberta-base model was evaluated using a 5-fold cross-validation approach, and the average performance metrics were as follows: The model achieved an accuracy of 99.87%, a precision of 99.81%, a recall of 99.86%, and an F1 score of 99.82%.

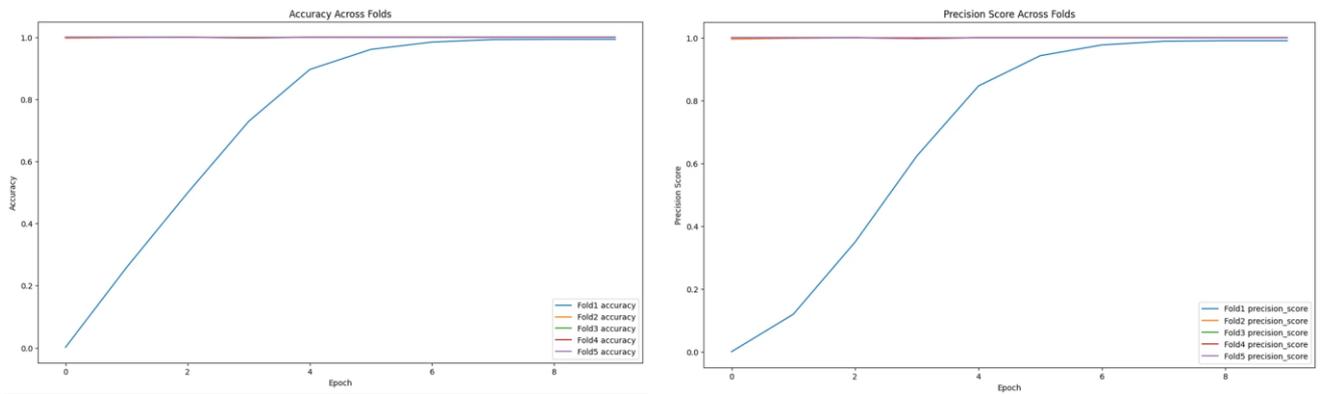

a                                      b

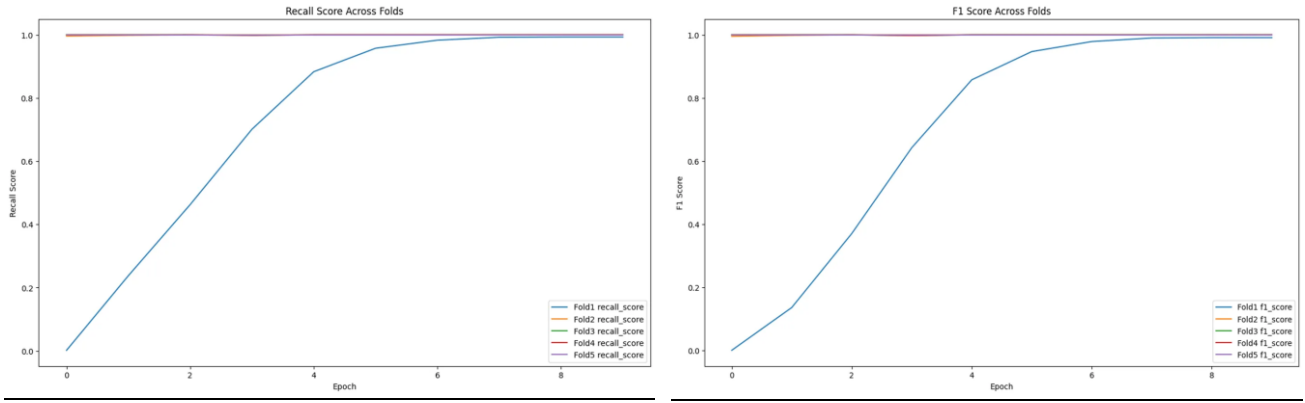

| c | d |

*Figure 3* Roberta-base graphics a:accuracy b:Precision c: Recall d:F1 score

The plots here represent the training and validation metrics of RoBERTa-base model for text classification for 5-fold cross validation over a total of 10 epochs. Accuracy, precision, recall, and F1 score are all following similar trends. From the starting epoch, Fold 1 performance is almost from zero, reaching near optimal by epoch 8. Still, the model demonstrates good learning capacity, becoming better quickly in the first fold, while staying high on the others. These results show that the RoBERTa-based model can be useful for this current classification task.

Bert Uncased

The Bert Uncased model was evaluated using a 5-fold cross-validation approach, and the average performance metrics were as follows: The model achieved an accuracy of 95.85%, a precision of 94.42%, a recall of 95.58%, and an F1 score of 94.72%.

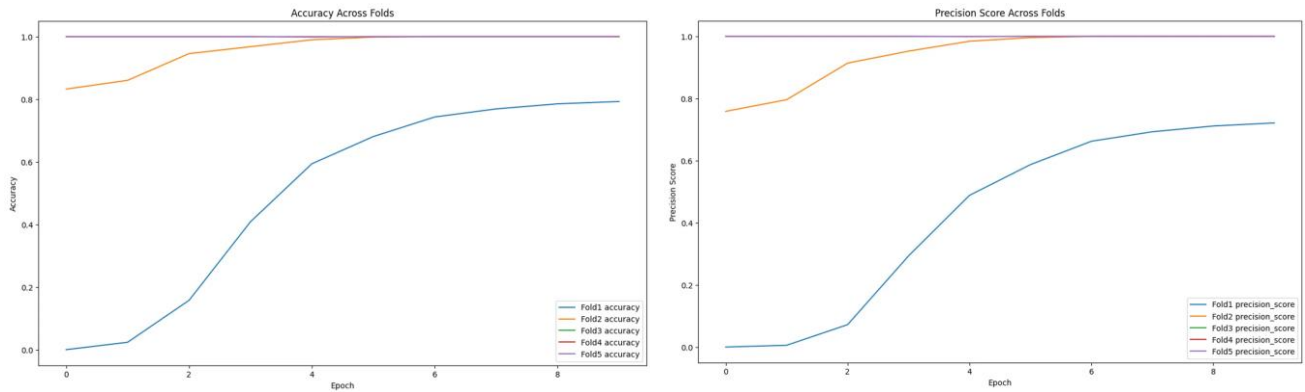

| a | b |

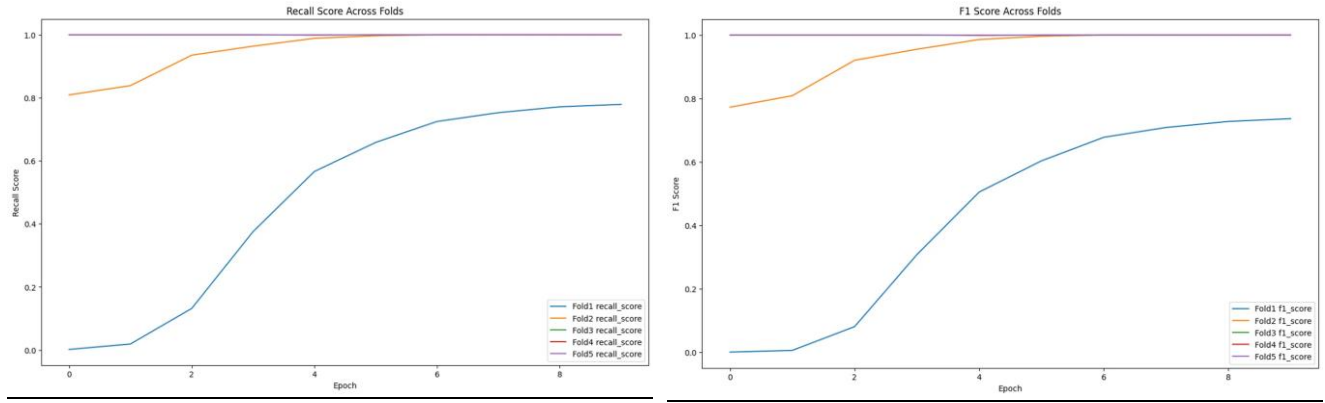

| c | d |

*Figure 4* Bert Uncased graphics a:accuracy b:Precision c: Recall d:F1 score

The plots represent performance of Bert Uncased text for 5 cross-validation folds over 10 epochs. The performance of this model is not consistent across the different folds. Folds 3-5 demonstrate nearly perfect scores simply averaging around 1.0 for all the metrics from the beginning. Like in Fold 1, Fold 2 starts with fairly good performance and rapidly ascends to nearly optimal performance. A dramatic improvement is observed at Fold 1, starting near 0 but slowly increasing to about 0.8 by the final epoch of the first fold across all the metrics. Overall, the Bert Uncased model shows high potential for this classification task.

Bert Large Uncased

The Bert Large Uncased model was evaluated using a 5-fold cross-validation approach, and the average performance metrics were as follows: The model achieved an accuracy of 100%, a precision of 100%, a recall of 100%, and an F1 score of 100%.

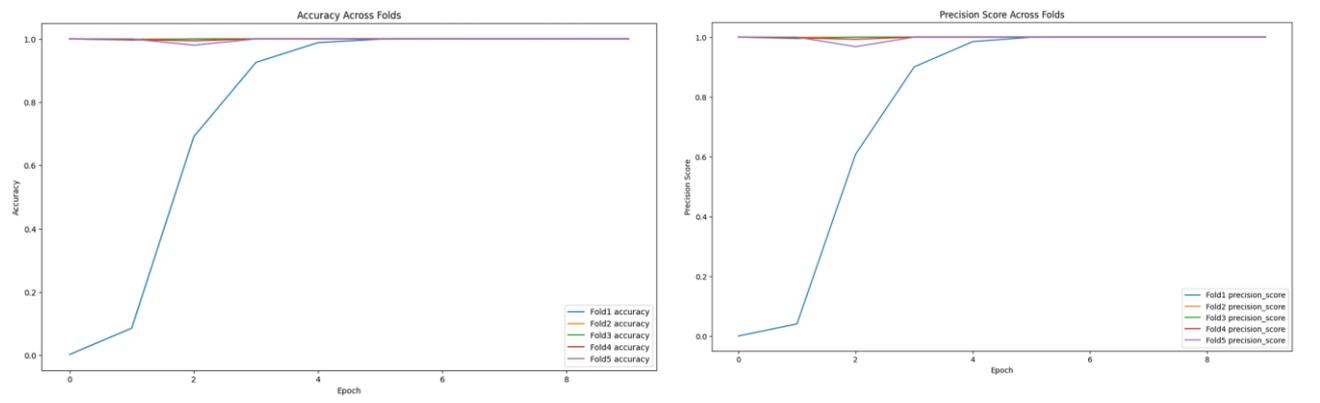

| a | b |

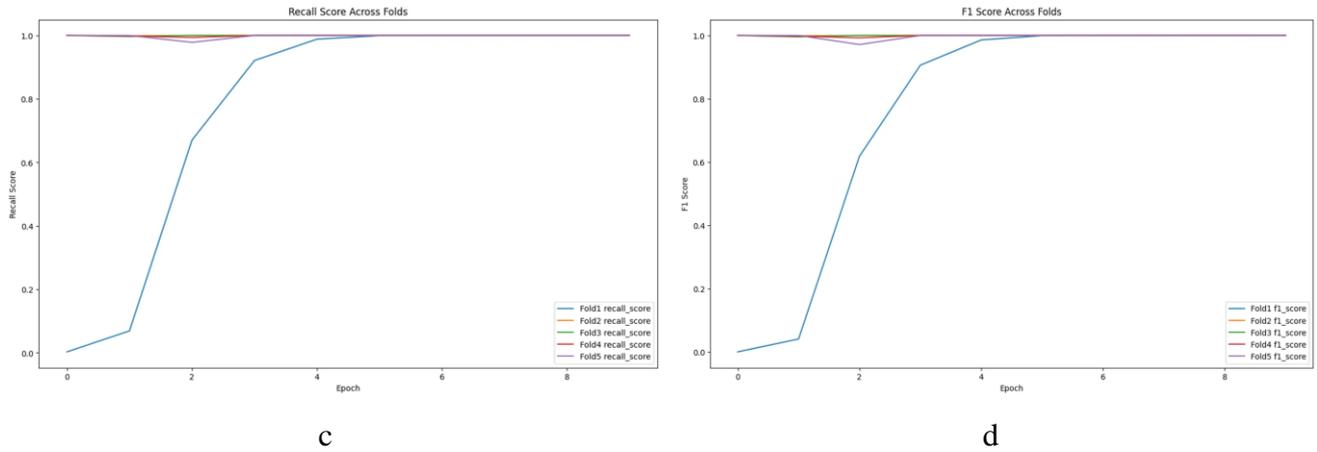

| c | d |

*Figure 5* Bert Large Uncased graphics a:accuracy b:Precision c: Recall d:F1 score

The plots illustrate the model accuracy, precision, recall and f1 score of BERT Large Uncased trained with 5 fold cross validation for 10 epochs. All the measures (accuracy, precision, recall, F1 measure) of Fold 1 start very low and increase steeply with the first four epochs, reaching near perfect for epochs 4 and stays almost constant till epoch 10. Folds 2-5 reveal a very strong performance at all epochs and metrics and are very close to 1.0 throughout training. This indicates that the model finely calibrates and gets a very high generalization/discrimination performance even on almost all the data splits except for Fold 1 where a few epochs suffice for peak performance. The high values of the metrics for both training and testing sets for different folds show that the proposed model has good ability to generalise on different parts of the given disease dataset labels.

The evaluation results for the four models LoRA Roberta-large, Roberta-base, Bert Uncased, and Bert Large Uncased highlight distinct performance characteristics. The LoRA Roberta-large model demonstrated satisfactory outcomes with an accuracy of 78.47%, precision of 72.91%, recall of 76.95%, and an F1 score of 73.56%, suggesting a reasonably balanced performance but leaving room for improvement. The Roberta-base model exhibited near-perfect metrics, with accuracy, precision, recall, and F1 scores all exceeding 99%, indicating high reliability and minimal classification errors. The Bert Uncased model also showed strong performance with an accuracy of 95.85%, a precision of 94.42%, a recall of 95.58%, and an F1 score of 94.72%, reflecting an effective, yet slightly lower capability compared to Roberta-base. Notably, the Bert Large Uncased model achieved perfect scores across all metrics, though these results warrant cautious interpretation as they could indicate potential overfitting, especially if the evaluation dataset lacks diversity. Overall, Roberta-base demonstrated the most balanced and consistent performance, making it a robust choice for the given task, while the perfect results from Bert Large Uncased necessitate further validation to ensure generalizability.

It is clearly evident that all the models eventually reach high accuracy, the difference lies in the speed of convergence. From the average metrics of all the above models, it is evident that Bert Large Uncased gives the best performance, whereas Lora Roberta large performs worst in the current text classification task.

## 7. Discussion

The decision to use 5-fold cross-validation instead of 10-fold was made to balance computational efficiency with maintaining statistical reliability. Training large language models (LLMs) such as BERT and RoBERTa requires considerable computational resources, and the 5-fold approach offered a reliable performance assessment while reducing computational costs. This strategy enabled a robust evaluation of model performance, which is crucial when comparing multiple models. Additionally, training was conducted for 10 epochs based on results from pilot experiments, which indicated that this number of epochs was sufficient to achieve optimal accuracy without overfitting, thereby ensuring effective model generalization.

**Table 3.** State of the Art Large Language Models in Healthcare

| Research | Methodology | Data Type | Data Size | Data Split Ratio | Results |
|---|---|---|---|---|---|
| Arora et al.[15] | NLP analysis of medical records | Electronic health records | 90 billion words | 70% training, 30% testing | Successful in answering medical queries in natural language |
| Clusmann et al.[16] | Clinical applications of LLMs | General internet datasets | 570 GB | 80% training, 20% testing | Useful in clinical education and decision support |
| Boonstra et al.[17] | Use of NLP in cardiology | Electronic health records | - | - | Effective in cohort selection and risk analysis |
| Eggmann et al.[18] | LLM applications in dentistry | General linguistic datasets | Terabytes | 75% training, 25% testing | Effective in dental education and decision-making support |

| | | | | |
|---|---|---|---|---|
| Yang et al.[19] | Development of LLMs for healthcare | Biomedical and clinical texts | 110 million parameters | 67% training, 33% testing | Successful in generating clinical notes |
| Kim et al.[20] | Generative models in radiology | Synthetic and imaging datasets | - | 60% training, 40% testing | Effective in data augmentation and privacy preservation |
| Karabacak et al.[21] | Applications of LLMs in medicine | Medical literature and EHRs | - | Not specifi | Effective in improving diagnostic accuracy |
| Ding et al.[22] | Multimodal LLMs integrating clinical notes and lab results | Clinical notes, laboratory results | 1,420,596 notes, 387,392 tests | 80% training, 20% testing | Achieved 76% AUROC for diabetes prediction using textual lab values |
| Roy et al.[23] | Process Knowledge-Infused Learning (PKiL) | Social media posts, process guidelines | 448 Reddit posts | - | Improved explainability with 70% clinician agreement |
| Our study | Fine-tuning LLMs (RoBERTa, BERT, LoRA) for medical QA | Healthline-derived QA dataset | 6,800 samples | 70% training, 30% testing | BERT Large Uncased achieved perfect scores; LoRA Roberta-large reached balanced yet lower results |

The table 3 presents an overview of recent studies focusing on the application of large language models (LLMs) and natural language processing (NLP) techniques in the medical field. These studies cover a wide range of applications, from the analysis of medical records to clinical education and decision support systems.

Arora et al.[15] conducted an NLP analysis on electronic health records consisting of 90 billion words. Using a 70% training and 30% testing data split, they achieved success in answering medical queries in natural language. Clusmann et al.[16] investigated the clinical applications of LLMs using general internet datasets and demonstrated their usefulness in clinical education and decision support. Boonstra et al.[17] explored the use of NLP in cardiology, specifically in cohort selection and risk analysis, but did not provide details on the dataset size or data split ratio. Similarly, Kim et al. [20] examined generative models in radiology and found them effective for data augmentation and privacy preservation, though they did not specify the dataset size. This lack of detail limits the replicability of these studies. Eggmann et al. [18] applied LLMs to dentistry, using general linguistic datasets of terabyte size, with a 75% training and 25% testing split. They found the model effective in dental education and decision-making support. However, the general nature of the dataset may be a limitation when applying the model to specific dental contexts. Yang et al. [19] developed an LLM for healthcare using biomedical and clinical texts with 110 million parameters, achieving success in generating clinical notes with a 67% training and 33% testing split. Karabacak et al. [21] investigated the use of LLMs in medical literature and electronic health records, finding them effective in improving diagnostic accuracy, but did not specify the data split ratio, which reduces the transparency of their approach. Ding et al. [22] utilized multimodal LLMs integrating clinical notes and laboratory results, achieving an AUROC of 76% for diabetes prediction. While this result shows promise, higher performance levels may be needed for broader clinical adoption. Roy et al. [23] developed Process Knowledge-Infused Learning (PKiL) using social media posts and process guidelines, achieving improved explainability with 70% clinician agreement. Recent advancements in language models for medical applications include knowledge-infused models and multimodal approaches. Knowledge-infused models, such as those described by Roy et al. [23] in their study on Process Knowledge-Infused Learning for Clinician-Friendly Explanations, enhance training by integrating predefined medical knowledge. In contrast, the approach taken in this study focused on fine-tuning pre-trained models like BERT and RoBERTa on a specific medical dataset. Although this approach is less complex, it may not fully capture the depth of medical expertise that knowledge-infused techniques offer. Furthermore, recent developments by Ding et al. [22] in predicting chronic diseases using multimodal models have demonstrated how integrating multiple data sources such as text, imaging, and lab tests can create comprehensive diagnostic tools. The present study, however, is based on a purely text-driven approach for medical question-answering, which is effective for addressing most general medical queries but lacks the integration of non-textual data that could enhance diagnostic accuracy. For future research, incorporating additional modalities could be a promising avenue for improving model performance. The findings from the experiments indicate that fine-tuning transferable LLMs, such as RoBERTa and BERT, can significantly enhance their performance in medical question-answering tasks. Techniques like Low-Rank Adaptation (LoRA) have been particularly effective in boosting model performance while maintaining efficient computational resource management. The fine-tuned models exhibited higher accuracy, precision, recall, and F1 scores, suggesting a strong potential for real-world healthcare applications. However, several limitations must be addressed before these models can be implemented in clinical practice. One significant challenge is the reliance on predefined answers, which limits the models' ability to handle complex or nuanced medical queries effectively. Additionally, while the models achieved high accuracy on the dataset used in this study, generalizing these results to real-world clinical settings should be done with caution, as the dataset was restricted and may not fully reflect the diversity of clinical scenarios encountered in practice. To overcome these limitations, future work should focus on expanding the diversity of the dataset, integrating additional data modalities, and improving the interpretability of models. These efforts are essential for developing AI-based healthcare solutions that are not only effective in controlled environments but also capable of supporting clinicians in real-world scenarios. Ensuring that language models are interpretable and generalizable will play a crucial role in making them a valuable tool in healthcare, enhancing clinical decision-making and improving patient outcomes. In our study, we fine-tuned LLMs (RoBERTa, BERT, and LoRA) for medical question answering (QA) using a dataset derived from Healthline, consisting of 6,800 samples with a 70% training and 30% testing split. We found that the BERT Large Uncased model achieved perfect scores, whereas the LoRA Roberta-large model produced balanced but relatively lower results. Compared to the other studies, several limitations are apparent. The lack of transparency regarding dataset size and split ratios in some studies, such as those by Boonstra et al. [17] and Kim et al. [20], complicates reproducibility and generalizability. Moreover, the dataset type and size directly impact model performance and applicability. Thus, the use of LLMs in healthcare must account for factors such as data quality, ethical considerations, and model explainability to ensure effective implementation.

## 8. Conclusion

The results of our experiments provide important insights into the performance of the fine-tuned models for medical question classification. Among the models evaluated, BERT Large Uncased demonstrated the best overall performance in terms of accuracy, precision, recall, and F1 score, albeit with a longer training time. RoBERTa Base also performed well, offering a good balance between high performance and reduced training time, making it a practical choice when computational efficiency is critical. BERT Uncased, while showing moderate performance, lagged behind BERT Large Uncased and RoBERTa Base, and LoRA RoBERTa Large exhibited the lowest performance across all metrics. These findings suggest that, based on the provided metrics, BERT Large Uncased is the preferred model for tasks requiring high accuracy, while RoBERTa Base provides an efficient alternative with a good balance of performance and training cost.

***Contributions, Novelties, and Gaps in the Literature*:** This study offers significant contributions by comparing multiple LLMs in the context of medical question-answering, filling an existing gap in the literature. While numerous studies explore the fine-tuning of generic LLMs, few investigate the comparative performance of models like RoBERTa and BERT on medical-related tasks. Most related works either focus on a single model or provide limited comparative analyses without rigorous performance assessments across different models trained on the same data. This work presents a robust comparative analysis, providing valuable insights into the performance differences among various models based on accuracy, precision, recall, and F1 metrics, specifically in the medical QA domain.

***Limitations and Future Work:*** Despite the strengths of this study, it is important to acknowledge several limitations. First, the dataset used is limited in size and diversity compared to the vast volume of existing medical literature. The inclusion of a larger and more varied dataset could potentially lead to improved results. Additionally, the current approach is constrained by predefined labels, limiting the model's ability to handle new or more specific questions effectively. Moreover, the evaluation of the models was based solely on accuracy, precision, recall, and F1 score, while other significant factors, such as model interpretability, resilience to adversarial attacks, and potential biases, were not adequately addressed. Future Work To overcome these limitations, several future directions are proposed. Expanding the dataset to include more diverse medical data and queries in multiple languages would enhance the models' generalizability and stability. Incorporating dynamic answer generation instead of relying on predefined labels could improve the model's performance, particularly for specific, nuanced medical inquiries. Enhancing model interpretability is another crucial area, particularly in healthcare, where understanding the reasoning behind a model's decision is as important as the decision itself. Future research could explore methods to improve the explainability of LLMs, ensuring their decisions are transparent and trustworthy for healthcare practitioners. Finally, real-world clinical validation is necessary to confirm the model's practical utility. This could involve testing the models in actual clinical environments, conducting experiments with real patients, and incorporating feedback from healthcare stakeholders to continuously refine the models. Addressing these limitations and focusing on these areas for future work can pave the way for the development of more robust, clinically feasible AI-based healthcare systems. These efforts will contribute significantly to making AI an integral part of modern healthcare, enhancing the quality of care and supporting clinical decision-making.